\documentclass[a4paper]{llncs}

\usepackage{amssymb}
\usepackage{amsmath}
\usepackage{graphicx}
\usepackage{multirow}
\usepackage{url}  

\newcommand{\keywords}[1]{\par\addvspace\baselineskip
\noindent\keywordname\enspace\ignorespaces#1}

\begin{document}

\mainmatter  

\title{Data Mining using Unguided Symbolic Regression on a Blast Furnace Dataset}
\titlerunning{Unguided Symbolic Regression on a Blast Furnace Dataset}

\author{Michael Kommenda\inst{1}\thanks{The final publication is available at \protect\url{http://link.springer.com/chapter/10.1007/978-3-642-20525-5_28}}
 \and Gabriel Kronberger\inst{1} \and Christoph Feilmayr\inst{2} \and Michael Affenzeller\inst{1}}
\authorrunning {Michael Kommenda, et al.}

\institute{
    Heuristic and Evolutionary Algorithms Laboratory\\
		School of Informatics, Communications and Media\\
    Upper Austria University of Applied Sciences, Campus Hagenberg\\
    Softwarepark 11, 4232 Hagenberg, Austria\\
    \email{\{michael.kommenda,gabriel.kronberger,michael.affenzeller\}@fh-hagenberg.at}\\
    \vspace{0.2cm}
    \and voestalpine Stahl GmbH, voestalpine-Stra\ss e 3, 4020 Linz, Austria,\\
    \email{\{christoph.feilmayr,\}@voestalpine.com}
   }

\maketitle

\begin{abstract}
In this paper a data mining approach for variable selection and
knowledge extraction from datasets is presented. The approach is based
on unguided symbolic regression (every variable present in the dataset
is treated as the target variable in multiple regression runs) and a
novel variable relevance metric for genetic programming. The relevance
of each input variable is calculated and a model approximating the
target variable is created. The genetic programming configurations
with different target variables are executed multiple times to reduce
stochastic effects and the aggregated results are displayed as a
variable interaction network. This interaction network highlights
important system components and implicit relations between the
variables. The whole approach is tested on a blast furnace dataset,
because of the complexity of the blast furnace and the many
interrelations between the variables. Finally the achieved results are
discussed with respect to existing knowledge about the blast furnace
process.  \keywords{Variable Selection, Genetic Programming, Data
  Mining, Blast Furnace}
\end{abstract}

\section{Introduction}
Data mining is the process of finding interesting patterns in large
datasets to gain knowledge about the data and the process it
originates from. This work concentrates on the identification of
relevant variables which is mainly referred to as variable or feature
selection (\cite{Guyon2003} provides a good overview about the
field). Usually a large set of variables is available in datasets to
model a given fact and it can be assumed that only a specific subset
of these variables is actually relevant. Although there are often no
details given on how variables are related, an identified set of
relevant variables is easy to understand and can already increase the
knowledge about the dataset considerably. However, determining the
subset of relevant variables is non-trivial especially if there are
non-linear or conditional relations. Implicit dependencies between
variables further hamper the identification of relevant variables as
this ultimately leads to multiple sets of different variables that are
equally possible.

In this paper genetic programming (GP) \cite{Koza1992}, a general
problem solving meta-heuristic, is used for data mining. GP is well
suited for data mining because it produces interpretable white box
models and automatically evolves the structure and parameters of the
model \cite{Koza1992}. In GP feature selection is implicit because
fitness-based selection makes models containing relevant variables
more likely to be included in the next generation. As a consequence,
references to relevant variables are more likely than references to
irrelevant ones. This implicit feature selection also removes
variables which are pairwise highly correlated but irrelevant to
describe a given relation. However, if pairwise correlated and
relevant variables exists in the dataset, GP does not recognize that
one of the variables can be removed and keeps both.

In this work symbolic regression analysis is executed multiple times
to reveal sets of relevant variables and to reduce stochastic
events. Additionally aggregated characteristics about the whole
algorithm run are used to extract information about the dataset,
instead of solely using the identified model. In section
\ref{sec-variable-relevance} a overview of metrics used to calculate
the variable relevance is given and a new frequency-based variable
relevance metric is proposed. Section \ref{sec-experiments} outlines
the experimental setup, the blast furnace dataset and the parameters
for the GP runs. Section \ref{sec-results} presents and discusses the
achieved results and section \ref{sec-conclusion} concludes the paper.

\section{Variable Relevance Metrics for GP}
\label{sec-variable-relevance}
Knowledge about the minimal set of input variables necessary to
describe a given dependent variable is often very valuable for domain
experts and can improve the understanding of the examined system. In
the case of linear models the relevance of variables can be detected
by shrinkage methods \cite{Hastie2009}. If genetic programming is used
for the analysis of relevant variables not only linear relations but,
based on the set of allowed symbols, also non-linear or conditional
impact factors can be detected. The extraction of the variable
relevance from GP runs is not straightforward and highly depends on
the metrics used to measure the variable importance.

Two variants to approximate the relevance of variables for genetic
programming have been described in \cite{Winkler2008}. Although both
metrics have been designed to measure population diversity they can be
used to estimate the variable relevance. The frequency-based approach
either uses the sum of variable references in all models or the number
of models referencing a variable. The second, impact-based metric uses
the information present in the variable to estimates its
relevance. The idea is to manipulate the dataset to remove the
variable for which the impact should be calculated (e.g., by replacing
all occurrences with the mean of the variable) and to measure the
response differences between the original model and the manipulated
one.

In \cite{Vladislavleva2010b} two different definitions of variable
relevance are proposed. The presence weighted variable importance
calculates the relative number of models, identified and manually
selected by one or multiple ParetoGP \cite{Smits2005} runs, which
reference this variable. The fitness-weighted variable importance
metric also uses the presence of variables in identified models, but
additionally takes the fitness of the identified models into account
\cite{Smits2006}. As the authors state this eliminates the need of
manually selecting models because the aggregated and weighted score of
irrelevant variables should be much smaller than the overall score of
relevant variables.

\subsection{Extension of Frequency-based Variable Relevance for GP}
The frequency-based variable relevance $\text{rel}_{\text{freq}}$ is
also based on the variable occurrence over multiple models but in
contrast to the other metrics the whole algorithm run is used to
calculate the variable relevance. The frequency of a variable $x_i$ in
a population of models is calculated by counting the references to
this variable over all models $m$ (Equation
\ref{equ-count},\ref{equ-frequ}). The frequency is afterwards
normalized by the total number of variable references in the
population (Equation \ref{equ_rel_frequ}) and the resulting
frequencies are averaged over all generations (Equation
\ref{equ_var_impact}).

\begin{equation}
\label{equ-count}
\text{CountRef}(x_i,m) = 
\begin{cases}
	1+ \sum_{b \in \text{Subtrees}(m)}^{ } \text{CountRef}(x_i,b) \text{ , } if \text{ Symbol}(m) = x_i\\
  0+ \sum_{b \in \text{Subtrees}(m)}^{ } \text{CountRef}(x_i,b) \text{ , } if \text{ Symbol}(m) \neq x_i\\
\end{cases}
\end{equation}

\begin{equation}
	\label{equ-frequ}
	\text{frequ}(x_i,\text{Pop})=  \sum_{m \in \text{Pop}}^{ } \text{CountRef}(x_i,m)
\end{equation}

\begin{equation}
	\label{equ_rel_frequ}
	\text{rel}_{\text{frequ}}(x_i,\text{Pop})= \frac{\text{freq}(x_i,\text{Pop)}}{\sum_{k=1}^{n} \text{freq}(x_k,\text{Pop})}
\end{equation}

\begin{equation}
	\label{equ_var_impact}
	\text{relevance}(x_i)= \frac{1}{G} \sum_{g=1}^{G} \text{rel}_{\text{freq}}(x_i,\text{Pop}_g)
\end{equation}

\begin{figure}
\centering
\includegraphics*[scale=0.7]{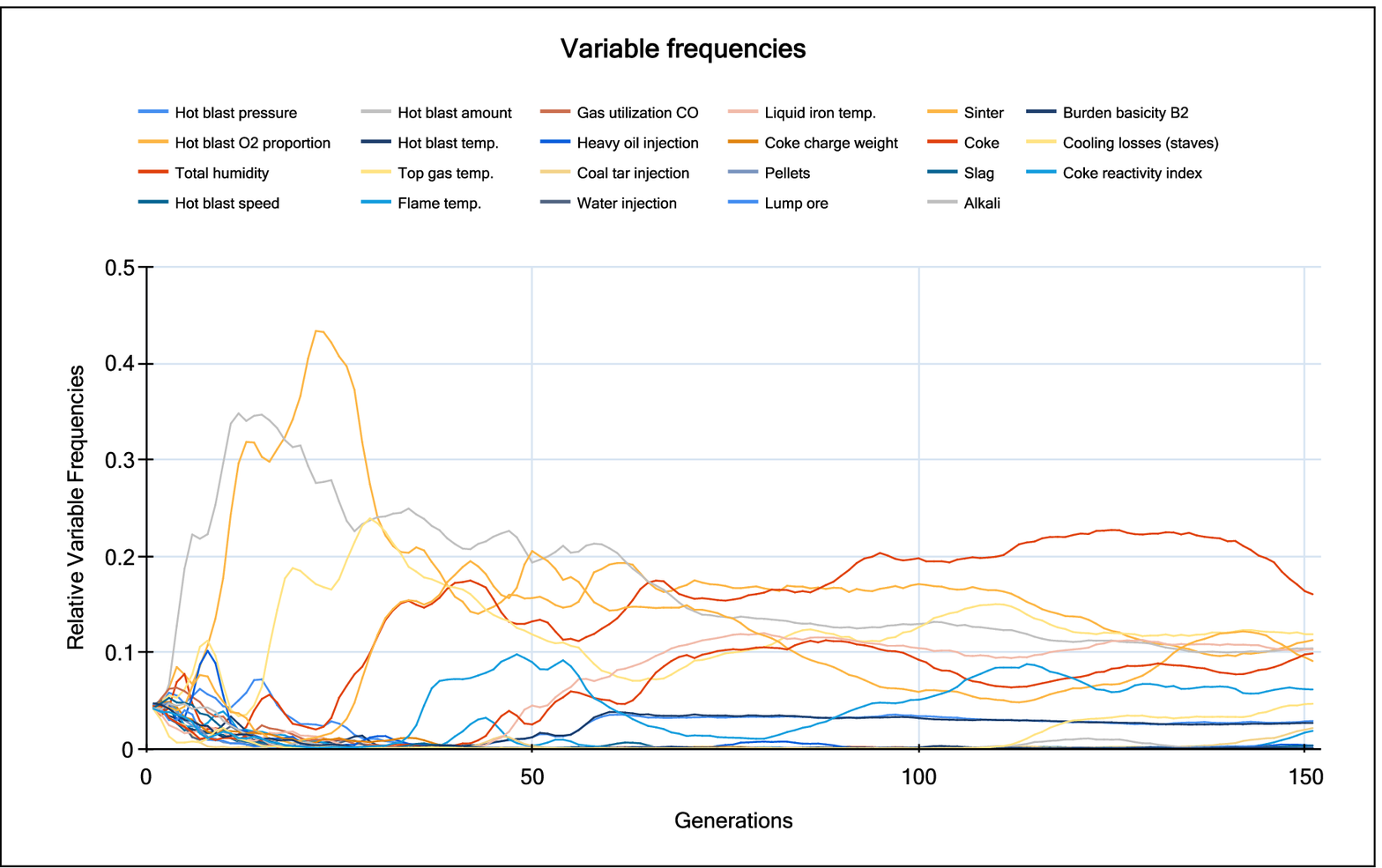}
\caption{Relative Variable Frequencies of one single GP run for the blast furnace dataset.}
\label{fig-variable-frequencies}
\end{figure}

Tracing the relative variable frequencies over the whole GP run and
visualizing the results is aimed to lead to insights into the dynamics
of the GP run itself. Figure \ref{fig-variable-frequencies} shows the
trajectories of relative variable frequency for the blast furnace
dataset described in section \ref{sec_dataset}. It can be already seen
that the relevance of variables varies during the GP run. In the
beginning two variables (the hot blast amount and the hot blast $O_2$
proportion) are used in most models, but after 100 generations the
total humidity overtops these two. The advantage of calculating the
variable relevance over the whole run instead of using only the last
generation is that the dynamic behavior of GP is taken into account.

Because of the non-deterministic nature of the GP process the
relevance of variables typically differs over multiple independent GP
runs. Implicit linear or non-linear dependencies between input
variables are another possible reason for these
differences. Therefore, the variable relevances of one single GP run
are not representative. It is desirable to analyze variable relevance
results over multiple GP runs in order to know which variables are
most likely necessary to explain the target variable and which
variables have a high relevance in single runs only by
chance. Therefore, all GP runs are executed multiple times and the
results are aggregate to minimize stochastic effects.

\section{Experiments}
\label{sec-experiments}
The frequency-based variable relevance metric and data mining approach
is tested on a complex industrial system. The general blast furnace
and the physical and chemical reactions occurring in the blast furnace
are quite well known. However, on a detailed level many of the
inter-relationships of different parameters and the occurrence of
fluctuations and unsteady behavior in the blast furnace are not
totally understood. Therefore, the knowledge about relevant variables
and accurate approximations of process variables are of special
importance and were calculated using repeated GP runs on the blast
furnace dataset.

\subsection{Blast Furnace Dataset}
\label{sec_dataset}
The blast furnace is the most common process to produce hot metal
globally. More than 60\% of the iron used for steel production is
produced in the blast furnace process \cite{schmoele2007}. The raw
materials for the production of hot metal enter the blast furnace via
two paths. At the top of the blast furnace ferrous oxides and coke are
charged in alternating layers. The ferrous oxides include sinter,
pellets and lump ore. Additionally feedstock to adjust the basicity is
also charged at the top of the blast furnace. In the lower area of the
blast furnace the hot blast (air, $1200\,^{\circ}\mathrm{C}$) and
reducing agents are injected through tuyeres. These reducing agents
include heavy oil, pulverized coal, coke oven or natural gas, coke tar
and waste plastic and are added to substitute coke. The products of
the blast furnace are liquid iron (hot metal) and the liquid byproduct
slag tapped at the bottom and blast furnace gas which is collected at
the top. For a more detailed description of the blast furnace process
see \cite{Strassburger1969}.

The basis of our analysis is a dataset containing hourly measurements
of a set of variables of the blast furnace listed in Table
\ref{table:blast-furnace-variables}. The dataset contains almost 5500
rows; rows 100--3800 are used for training and rows 3800--5400 for
testing. Only the first half of the training set (rows 100--1949) is
used to determine the accuracy of a model. The other half of the
training set (rows 1950--3800) is used for validation and selection of
the final model. The dataset cannot be shuffled because the
observations are measured over time and the nature of the process is
implicitly dynamic.

\begin{table}
\centering
\small
\begin{tabular}{|l|ll|}
\hline
Group & \multicolumn{2}{c|}{Variables}  \\
\hline
\multirow{3}{*}{Hot blast} & pressure & amount\\
& $O_2$ proportion & speed\\
& temperature & total humidity\\
\hline
\multirow{2}{*}{Tuyere Injection} & amount of heavy oil & amount of water\\
& amount of coal tar & \\
\hline
\multirow{4}{*}{Charging} & coke charge weight & amount of sinter \\
& amount of pellets & amount of coke \\
& amount of lump ore & burden basicity B2\\
& coke reactivity index & \\ 
\hline
\multirow{3}{*}{Tapping} & hot metal temperature & amount of slag \\
& amount of alkali & \\
\hline
Blast furnace top gas & temperature & gas utilization CO \\
\hline
Process parameters & melting rate & cooling losses (staves)\\
\hline
\end{tabular}
\caption{Variables included in the blast furnace dataset.}
\label{table:blast-furnace-variables}
\end{table}

\subsection{Algorithmic Settings}
Unguided symbolic regression treats each of the variables listed in
Table \ref{table:blast-furnace-variables} as the target variable in
one GP configuration and all remaining variables are allowed as input
variables. This leads to 23 different configurations, one for each
target variable. For each configuration 30 independent runs have been
executed on a multi processor blade system to reduce stochastic
effects. Table \ref{table:gp-parameters-blast-furnace} lists the
algorithm parameters for the different GP configurations. The
resulting model of the GP run is that one with the largest $R^2$ on
the validation set and gets linearly scaled \cite{Keijzer2004} to fit
the location and scale of the target variables. The approach described
in this contribution was implemented and tested in the open source
framework HeuristicLab \cite{Wagner2009}.

\begin{table}
\centering
\small
\begin{tabular}{|l|l|}
\hline
Parameter & Value\\
\hline
Population size & 1000\\
Max. generations & 150\\
Parent selection & Tournament (group size =7)\\
Replacement & 1-Elitism \\
Initialization & PTC2 \cite{luke:2000:2ftcaGP} \\ 
Crossover & Sub-tree-swapping\\
Mutation rate & 15\%\\
Mutation operators & One-point and Sub-tree replacement\\
Tree constraints & Max. expression size = 100\\
& Max. expression depth = 10\\
Model selection & Best on validation\\
Stopping criterion & Max. generations reached\\
Fitness function & $R^2$ (maximization)\\
Function set & +,-,*,/,avg,log,exp\\
Terminal set & constants, variable \\
\hline
\end{tabular}
\caption{Genetic programming parameters for the blast furnace dataset.}
\label{table:gp-parameters-blast-furnace}
\end{table}

\section{Results}
\label{sec-results}
\begin{figure}
\centering
\includegraphics[height=8cm]{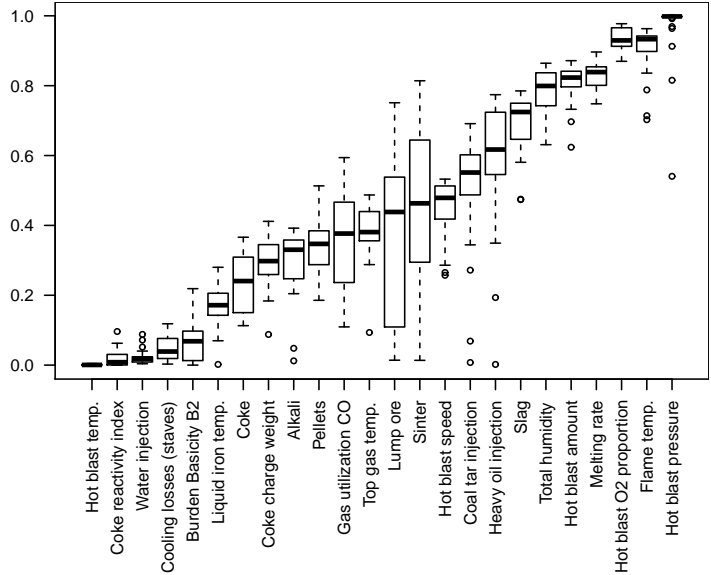}
\caption{Box-plot of $R^2$ value on the test set of models for the blast furnace dataset.}
\label{figure:blast-furnace-boxplot}
\end{figure}

A box plot of the model accuracies ($R^2$) over 30 independent runs
for each target variable of the blast furnace dataset is shown in
Figure \ref{figure:blast-furnace-boxplot}. The $R^2$ values are
calculated from the predictions of the best model (selected on the
validation set) on the test set for each run. Whiskers indicate four
times the interquartile range, values outside of that range are
indicated by small circles in the box-plot. Almost all models for the
hot blast pressure result in a perfect approximation ($R^2 \approx
1.0$). Very good approximations are also possible for the $O_2$
proportion of the hot blast and for the flame temperature. On the
other hand the hot blast temperature, the coke reactivity index and
the amount of water injected through tuyeres cannot be modeled
accurately using symbolic regression.

\subsection{Variable Interaction Network}
\begin{figure}
\centering
\includegraphics[width=9cm]{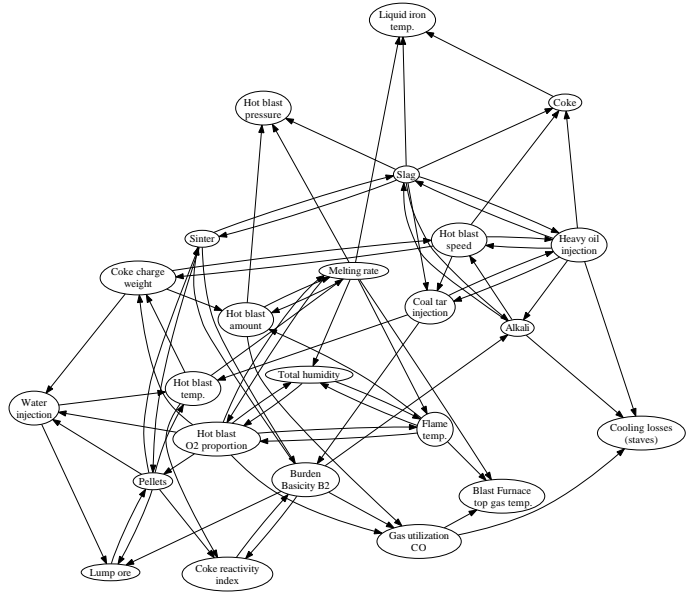}
\caption{Relationships of blast furnace variables identified with unguided symbolic regression.}
\label{figure:blast-furnace-network}
\end{figure}

The variable interaction network obtained from the GP runs is shown in
Figure \ref{figure:blast-furnace-network}. For each target variable
the three most relevant input variables are indicated by an arrow
pointing to the target variable. Arrows in both directions are an
indication that the pair of variables is strongly related; the value
of the first variable is needed to approximate the value of the second
variable and vice versa. Variables that have many outgoing arrows play
a central role in the process and can be used to approximate many
other variables. In the blast furnace network central variables are
the melting rate, the amount of slag, the amount of injected heavy
oil, the amount of pellets, and the hot blast speed and its $O_2$
proportion. The unfiltered variable interaction network must be
interpreted in combination with the box plot in Figure
\ref{figure:blast-furnace-boxplot} because the significance (not in
the statistical sense) of arrows pointing to variables which cannot be
approximated accurately is rather low (e.g., the connection between
the coke reactivity index and the burden basicity B2).

\subsection{Detailed Results}
The variable interaction network for the blast furnace process
provides a good overview of the blast furnace process. Exemplary the
influence factors obtained by unguided symbolic regression on the
melting rate are analyzed and compared to the influences known by
domain experts.

The melting rate is primarily a result of the absolute amount of $O_2$
injected into the furnace and is also related to the efficiency of the
furnace. A crude approximation for the melting rate is
\begin{equation}
\frac{\text{Total amount of } O_2}{[220\ldots 245]}
\end{equation}
When the furnace is working properly the melting rate is higher
($O_2/220$), when the furnace is working inefficiently the melting
rate decreases ($O_2/245$) and high cooling losses can be
observed. Additional factors that are known to affect the melting rate
are the burden composition and the amount of slag. The identified
models show a strong relation of the melting rate with the hot blast
parameters (data not shown). The melting rate is used in models for
the hot blast parameters: pressure, $O_2$-proportion, amount, and the
total humidity which is largely determined by the hot blast. In return
the hot blast parameters play an important role in the model for the
melting rate.

Equation \ref{eqn:melting-rate} shows a model for the melting rate
with a rather high squared correlation coefficient of 0.89 that has
been further simplified by omitting uninfluential terms and manual
pruning. The generated model \ref{eqn:melting-rate} (constants $c_i,
i=1..8$ are omitted for better readability) also indicates the known
relation of the melting rate and the amount of $O_2$. Additionally the
cooling losses, the amount of lump ore and the gas utilization of CO
have been identified as factors connected to the melting rate.

\begin{align}
\begin{split}
\text{Melting rate} = & {} \log(c_0\times \text{Temp}_\text{HB} \times O_2\text{-prop}_\text{HB} \times \\
& (c_1\text{Cool. loss} + c_2\text{Amount}_\text{HB} + c_3) + c_4 \times \text{Gas util}_{CO} \\
& \times (c_5\text{Lump ore} + c_6) \times (c_7\text{Amount}_\text{HB} + c_8) )
 \end{split}
\label{eqn:melting-rate}
\end{align}

\section{Conclusion}
\label{sec-conclusion}
Many variables in the blast furnace process are implicitly related,
either because of underlying physical relations or because of the
external control of blast furnace parameters. Examples for variables
with implicit relations to other variables are the flame temperature
or the hot blast parameters. Usually such implicit relations are not
known a-priori in data-based modeling scenarios but could be extracted
from the variable relevance information collected from multiple GP
runs.

Using an unguided symbolic regression data mining approach several
models have been identified that approximate the observed values in
the blast furnace process rather accurately. In some cases the
data-based models approximate known underlying physical relations, but
in general the statistical models produced by the data mining approach
do not match the physical models perfectly. A possible enhancement
could be the usage of physical units in the GP process to evolve
physically correct models.

Currently the variable relevance information is used to determine the
necessary variable set to model the target variable. The experiments
also lead to a number of models describing several components of the
blast furnace. The generated models can be used to extract information
about implicit relations in the dataset to further reduce and
disambiguate the set of relevant input variables. Additionally the
information about relations between input variables can be used to
manually transform symbolic regression models to lower the number of
alternative representation of the same causal relationship. However,
the implementation of software that uses such models of implicit
relations or manually declared a-priori knowledge intelligently, to
simplify symbolic regression models, or to provide alternative
semantically equivalent representations of symbolic regression models,
is left for future work.

\subsubsection*{Acknowledgments} 
This research work was done within the Josef Ressel-center for
heuristic optimization ``Heureka!'' at the Upper Austria University of
Applied Sciences, Campus Hagenberg and is supported by the Austrian
Research Promotion Agency (FFG).

\clearpage
\bibliographystyle{splncs03}
\bibliography{Kommenda}

\end{document}